\documentclass[10pt,twocolumn,letterpaper]{article}

\usepackage[pagenumbers]{iccv} 

\usepackage{multirow}
\usepackage{pifont}

\definecolor{iccvblue}{rgb}{0.21,0.49,0.74}
\usepackage[pagebackref,breaklinks,colorlinks,allcolors=iccvblue]{hyperref}
\usepackage[accsupp]{axessibility}

\def\paperID{7580} 
\def\confName{ICCV}
\def\confYear{2025}

\title{Cross-modal Ship Re-Identification via Optical and SAR Imagery: A Novel Dataset and Method}

\author{Han Wang \textsuperscript{1,2} \quad
Shengyang Li \textsuperscript{1,2}\thanks{Corresponding author} \quad
Jian Yang \textsuperscript{1,2} \quad
Yuxuan Liu \textsuperscript{1,2} \quad
Yixuan Lv \textsuperscript{1} \quad
Zhuang Zhou \textsuperscript{1} \quad \\
\textsuperscript{1}Technology and Engineering Center for Space Utilization, Chinese Academy of Sciences \\
\textsuperscript{2}University of Chinese Academy of Sciences\\
}

\begin{document}
\maketitle
\begin{abstract}
Detecting and tracking ground objects using earth observation imagery remains a significant challenge in the field of remote sensing.
Continuous maritime ship tracking is crucial for applications such as maritime search and rescue, law enforcement, and shipping analysis.
However, most current ship tracking methods rely on geostationary satellites or video satellites.
The former offer low resolution and are susceptible to weather conditions, while the latter have short filming durations and limited coverage areas, making them less suitable for the real-world requirements of ship tracking.
To address these limitations, we present the Hybrid Optical and Synthetic Aperture Radar (SAR) Ship Re-Identification Dataset (HOSS ReID dataset), designed to evaluate the effectiveness of ship tracking using low-Earth orbit constellations of optical and SAR sensors.
This approach ensures shorter re-imaging cycles and enables all-weather tracking.
HOSS ReID dataset includes images of the same ship captured over extended periods under diverse conditions, using different satellites of different modalities at varying times and angles.
Furthermore, we propose a baseline method for cross-modal ship re-identification, TransOSS, which is built on the Vision Transformer architecture.
It refines the patch embedding structure to better accommodate cross-modal tasks, incorporates additional embeddings to introduce more reference information, and employs contrastive learning to pre-train on large-scale optical-SAR image pairs, ensuring the model's ability to extract modality-invariant features.
Our dataset and baseline method are publicly available on \href{https://github.com/Alioth2000/Hoss-ReID}{https://github.com/Alioth2000/Hoss-ReID}.
\end{abstract}
    
\section{Introduction}
\label{sec:intro}

Object tracking in remote sensing involves detecting and associating objects across image sequence to obtain their trajectories.
Most existing ship tracking methods rely on geostationary or video satellites, but they face significant limitations.
While geostationary remote sensing satellites \cite{ref1, ref2, ref3, ref4} provide high temporal resolution and extensive coverage, their spatial resolution often falls short for accurate ship identity recognition, leading to potential misjudgments.
High-resolution video satellites can effectively track ships in port and ocean scenes \cite{ref5, ref7, ref8, ref20, ref21}, but they have limited coverage and short tracking duration, with each video lasting only 90 to 180 seconds \cite{ref9}.
As a result, the feasibility of tracking that relies exclusively on these satellites is considerably constrained.

\begin{figure}[t]
\centering
\includegraphics[width=3in]{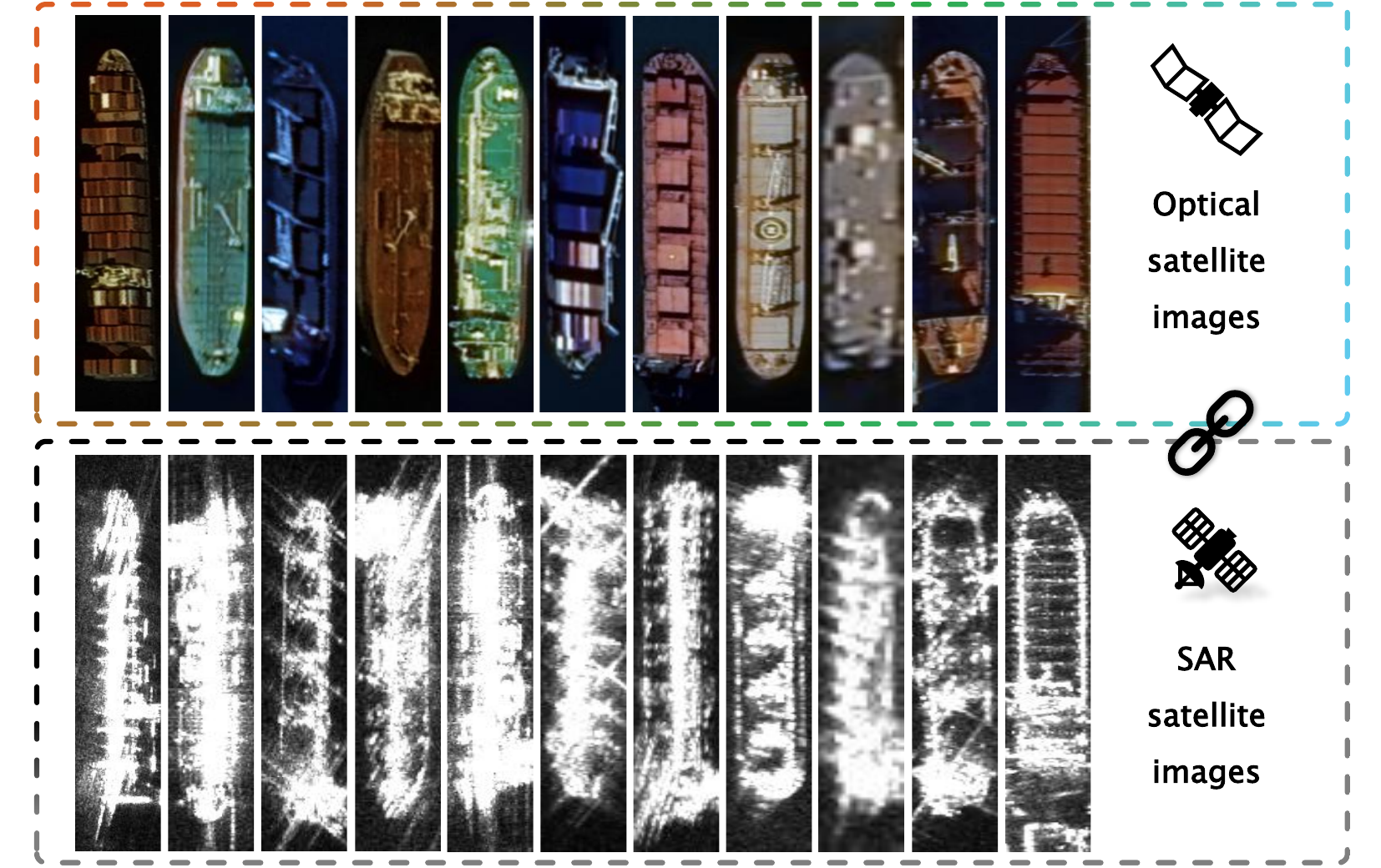}
\hfil
\caption{Examples from the HOSS ReID dataset. Images in the same column depict the same ship captured by different modalities of satellites, from various orbits, at different times. As shown in this figure, there is a significant difference in the imaging of the same object under optical and SAR modalities.\vspace{-3ex}}
\label{Dataset}
\end{figure}
In contrast, utilizing low Earth orbit (LEO) remote sensing constellations offers a more viable solution.
These constellations benefit from lower launch costs, a larger number of satellites, shorter revisit cycles, and higher spatial resolution, enhancing the likelihood of capturing objects and obtaining detailed features.
By leveraging the abundance and rapid revisit capabilities of LEO satellites, continuous imaging can be maintained, reducing the risk of losing sight of objects and facilitating long-term trajectory acquisition.
To effectively match objects across the vast number of captured images, robust re-identification (ReID) methods are essential.

\begin{figure}[t]
\centering
\includegraphics[width=3.2in]{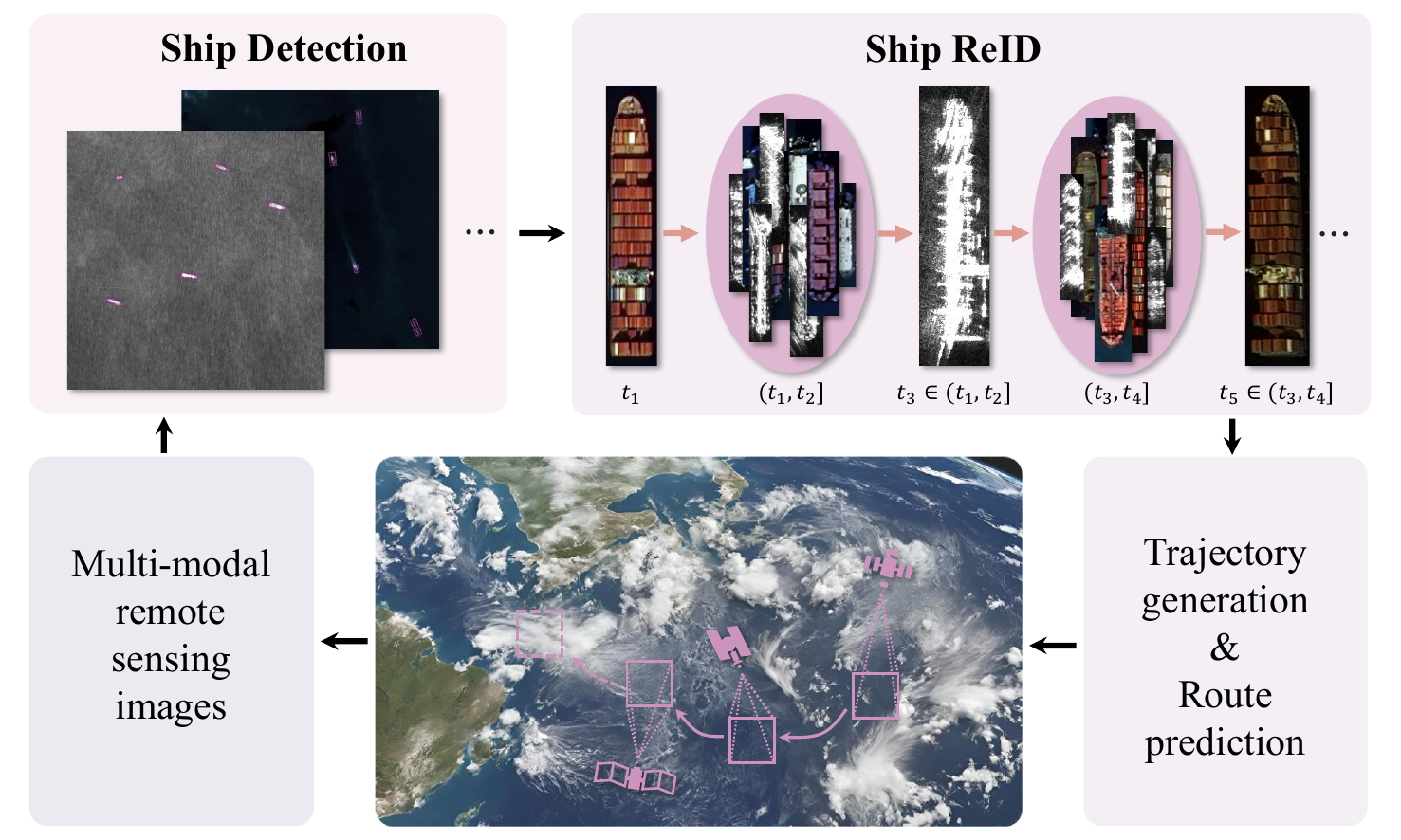}
\hfil
\caption{Schematic diagram of ship tracking method based on multimodal LEO constellations. By performing ReID on the ship of interest through remote sensing ship detection at different times and locations, the spatiotemporal trajectory of the ship can be established. This enables predicting the ship's trajectory to guide subsequent satellite imaging.\vspace{-3ex}}
\label{Shot}
\end{figure}

We propose to address the ship tracking problem from the perspective of constructing an integrated detection-ReID-trajectory generation pipeline, as illustrated in Figure \ref{Shot}.
While existing ship detection datasets and methods are well-established \cite{ref33, ref34, ref35, ref36, ref37, ref38}, our work focuses on ship ReID.
Object ReID seeks to associate specific objects across images captured in different scenes and by various cameras, with common applications including pedestrian and vehicle ReID.
This paper extends the concept of ReID to cross-modal ship ReID, focusing on linking ships in remote sensing images to facilitate tracking and trajectory estimation.
Once a satellite captures an interesting ship for the first time, the cross-modal ReID method can autonomously match the same object from a vast number of subsequent ship images captured at different locations and times.
This method leverages image data exclusively to achieve precise ship identification, which is particularly valuable for non-cooperative targets tracking.
Moreover, it can leverage the extensive constellation of LEO remote sensing satellites for continuous imaging, providing greater redundancy and extended tracking durations.
This capability is of paramount importance in critical fields such as rescue operations and law enforcement.

Optical remote sensing is highly vulnerable to weather conditions and is unable to detect or track objects at night.
In contrast, synthetic aperture radar (SAR) imaging overcomes these limitations by providing all-weather, day-and-night Earth observation capabilities.
However, the number of SAR satellites remains significantly lower than that of optical imaging satellites.
To minimize revisit times, integrating optical and SAR satellites is essential, enabling continuous maritime object imaging through a relay between optical and SAR satellites.
Despite this need, no existing cross-modal object tracking or ReID dataset or method specifically addresses the challenges of optical and SAR image object association.

For the reasons mentioned above, we present the Hybrid Optical and SAR Ship Re-Identification Dataset (HOSS ReID dataset), specifically designed to evaluate the effectiveness of ship tracking using LEO constellations of optical and SAR sensors.
Given the lack of suitable existing datasets, we create HOSS ReID dataset by acquiring new imagery.
We utilize the Jilin-1 constellation \cite{ref15} for optical images and the TY-MINISAR constellation \cite{ref16} for SAR images.
Some examples of the dataset are displayed in Figure \ref{Dataset}.

Additionally, we propose a cross-modal ship ReID method named TransOSS.
Unlike the common visible-infrared ReID tasks in pedestrian and vehicle ReID \cite{ref10, ref11, ref12, ref13, ref14}, optical-SAR ReID involves fundamentally different imaging mechanisms.
Optical imaging is passive, while SAR actively emits electromagnetic waves and receives echoes (typically employing centimeter-scale waves, which have wavelengths significantly longer than those of visible light and infrared radiation), capturing distinct scattering characteristics that produce markedly different imaging effects compared to visible light.
This makes optical-SAR ReID significantly more challenging than visible-infrared ReID.

To address the challenges of cross-modality ReID, we design an efficient architecture based on the Vision Transformer (ViT) \cite{ref17}.
This architecture employs a dual-head patch embedding mechanism, which separately embeds optical and SAR images, along with learnable modality information embeddings.
These modifications are aimed at bridging the substantial modality gap between optical and SAR images, suppressing irrelevant feature information, and guiding the model to map the images into a shared feature space.
Additionally, it incorporates a modality-shared transformer encoder, which simplifies training and enhances the extraction of common features.
Given that ship size can be directly measured from remote sensing images, we introduce a ship size embedding to mitigate the loss of size information caused by image resizing.
To further improve cross-modal learning, we develop a contrastive learning-based pretraining method using large-scale optical-SAR image pairs, enabling the model to effectively learn shared features across both modalities.

In summary, our main contributions are as follows:

\begin{itemize}
\item{We propose a technical approach for ship tracking through cross-modal ReID, enabling precise long-term and wide-area object monitoring via multiple LEO constellations.}

\item{We develop the first hybrid optical and SAR ship ReID dataset, collecting images of the same ship across different modalities, from various satellites, at different times, and from multiple angles.}

\item{We propose a novel cross-modal ship ReID method that incorporates a dual-head tokenizer and a modality-shared transformer encoder as the core architecture.}

\item{We introduce modality information embeddings and ship size embeddings, coupled with a two-stage training strategy. The model is pre-trained on a large-scale optical-SAR image pair dataset using contrastive learning, establishing a robust baseline performance.}
\end{itemize}

\section{Related Work}
\label{sec:related}

In this section, we begin by reviewing the current state of research on remote sensing ship tracking, followed by a summary of existing studies on cross-modal ReID.
\medskip

\noindent
{\bf Remote sensing ship tracking.}
Several existing datasets and algorithms have been developed for ship tracking, with some utilizing geostationary orbit (GEO) satellite image sequences.
The Automatic Identification System (AIS) is frequently employed to aid in tracking vessels \cite{ref18, ref19}.
For example, Yao \textit{et al.} \cite{ref3} used rational polynomial coefficients in combination with AIS data for dynamic correction, followed by an improved multiple hypothesis tracking algorithm with amplitude information to track ships.
Liu \textit{et al.} \cite{ref1} proposed a trajectory-level data fusion algorithm between GEO optical satellite imagery and AIS, enhancing vessel localization accuracy.
However, these algorithms are limited to tracking vessels with active AIS transponders, making it difficult to track non-cooperative targets.
Yu \textit{et al.} \cite{ref2} introduced a method for moving ship detection and tracking based on visual saliency, employing a joint probabilistic data association approach for data correlation and multi-object tracking across multiple frames.
Li \textit{et al.} \cite{ref4} proposed a novel multi-stage supervised network and a joint tracking method based on a low-frame-rate tracking criterion to minimize trajectory breaks.
While these GEO-based approaches enable ship detection and tracking, their low spatial resolution hinders precise ship identification and makes them highly susceptible to limitations caused by lighting and weather conditions.

Video satellites have also been employed for ship tracking.
Li \textit{et al.} \cite{ref21} released single-object tracking datasets for ships using satellite video, while Yin \textit{et al.} \cite{ref20} and Li \textit{et al.} \cite{ref8} introduced multi-object tracking datasets.
These video satellites typically offer a ground sample distance (GSD) of about 1 meter, with frame rates ranging from 5 to 30 frames per second, and capture footage lasting between 60 to 120 seconds \cite{ref9, ref21}.
Researchers have developed specialized methods tailored to these datasets, achieving precise single-object or multi-object ship tracking \cite{ref22, ref23, ref24, ref25}.
The main technical challenge lies in ship detection, as tracking is relatively straightforward but limited to short durations (tens of seconds) due to satellite motion.
While these methods reveal valuable dynamic attributes like course and speed, long-term tracking requires integrating other constellations and employing robust identity recognition and data association techniques.
Lang \textit{et al.} \cite{ref53} made initial attempts at ship tracking using SAR constellations, employing a detection-matching-tracking strategy that demonstrated the method's feasibility.
However, their work was limited by insufficient data and lacked further experimental validation.
\medskip

\noindent
{\bf Cross-modal ReID.}
Cross-modal ReID is commonly employed when RGB images alone are insufficient, and additional modalities, such as infrared or sketches, can provide complementary information.
Among these, infrared-visible pedestrian ReID has seen the most significant progress.
Several cross-modal pedestrian ReID datasets have been introduced, including SYSU-MM01 \cite{ref26} and RegDB \cite{ref55}, which have become widely accepted as standard benchmarks in the field.
Li \textit{et al.} \cite{ref13} proposed a method to bridge the gap between RGB and infrared images by embedding RGB images into an X-modality space using a lightweight network architecture.
Wu \textit{et al.} proposed MPANet \cite{ref54}, incorporating a modality alleviation module and a pattern alignment module to discover cross-modal nuances in different patterns.
In another study, Park \textit{et al.} \cite{ref28} addressed cross-modal discrepancies by leveraging dense correspondences between images of individuals across different modalities.
Further advancing this area, Wu \textit{et al.} \cite{ref29} proposed an unsupervised method based on progressive graph matching and cross contrastive learning to extract modality-invariant features, while Zhang \textit{et al.} \cite{ref30} introduced an embedding space enhancement network designed to generate more diverse embeddings, thus providing more informative feature representations.
In a novel approach, Li \textit{et al.} \cite{ref31} utilized a pre-trained large model as an encoder, facilitating effective multimodal ReID across RGB, infrared, sketch, and textual modalities without the need for additional fine-tuning.
Compared to infrared and visible images, optical and SAR images differ significantly, making it difficult for existing methods to suppress cross-modal redundancy, thus limiting their effectiveness in optical-SAR ReID tasks.

\section{HOSS ReID Dataset}
\label{sec:dataset}

We present, for the first time, a cross-modal ship ReID dataset called HOSS ReID, specifically designed for ship tracking, with the goal of addressing the challenges of all-time, all-weather and wide-area ship tracking. 

\begin{figure}[t]
\centering
\includegraphics[width=3.2in]{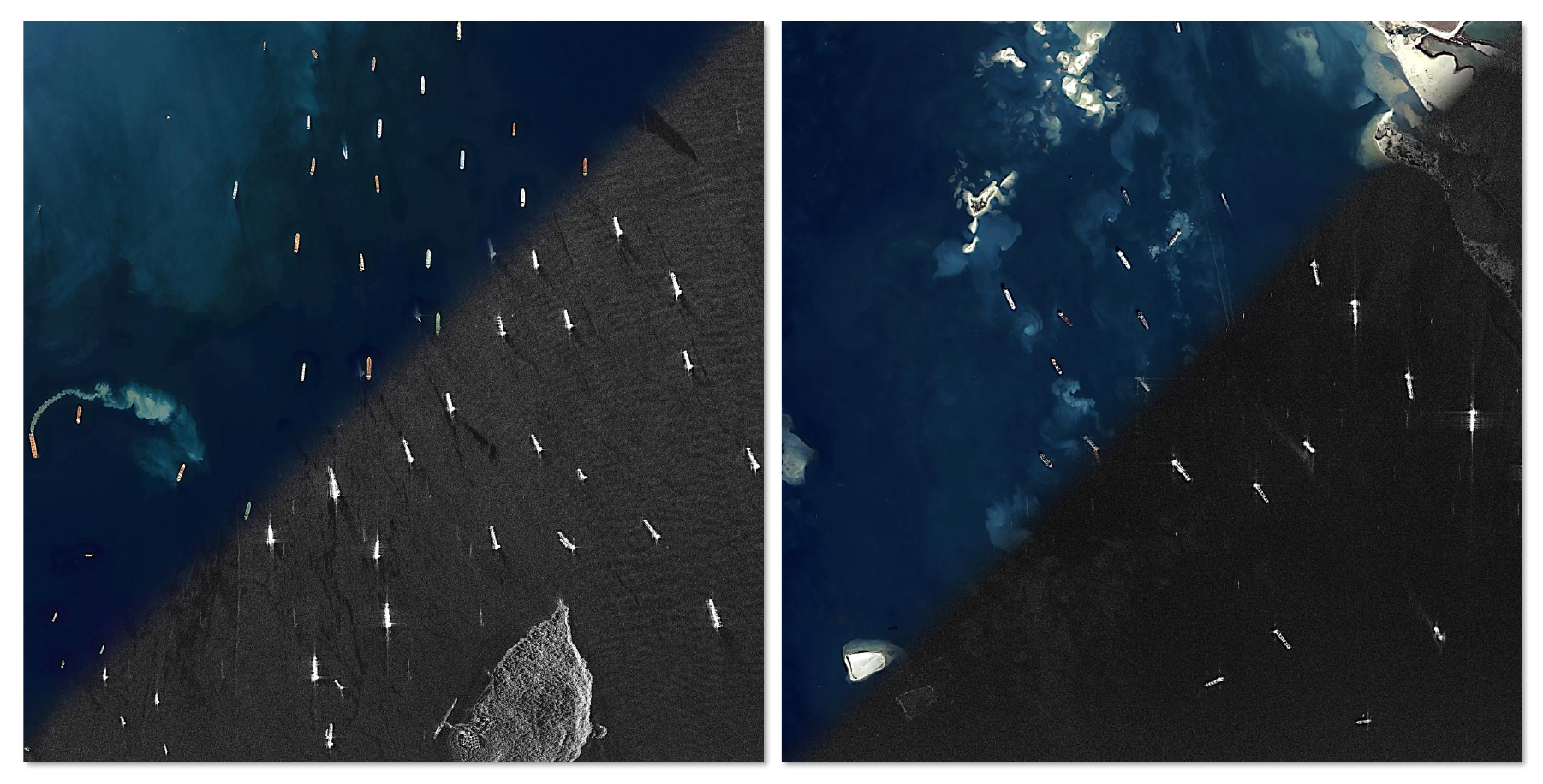}
\hfil
\caption{Some examples of images we have taken using LEO optical and SAR constellations. The left and right images depict the Panama Canal and the Suez Canal, respectively. The upper-left sections show optical images, while the lower-right sections display SAR images. We utilize multiple satellites to capture images within a short time span, allowing us to obtain multimodal imagery of the same ships.\vspace{-2.5ex}}
\label{Scenes}
\end{figure}

\subsection{Dataset Collection}
Given the stringent requirements for creating this dataset, publicly accessible data is nearly nonexistent, we opted to acquire the imagery from scratch through programmed satellite imaging.
However, this approach faces several challenges:
(1) the high cost of satellite tasking limits imaging opportunities;
(2) the mobility of ships hinders repeated observations;
(3) imaging conditions are constrained by weather, lighting, and satellite transit times, necessitating precise timing to capture multimodal sequences.

To mitigate these challenges, we implement several strategies:
(1) We prioritize ports, canals, and similar locations as primary imaging sites, enabling each capture to include a high density of ships;
(2) We minimize the interval between imaging sessions and concentrate on capturing ships that are anchored, thereby reducing the impact of vessel movement, where the ship's motion or precise location do not adversely affect the imaging results, and makes no difference for the ship ReID task..
Some examples of the images we captured are shown in Figure \ref{Scenes}.

\subsection{Dataset Description}
The optical images have a GSD of 0.75 meters, while the SAR images offer a GSD of 1 meter.
The image has undergone geometric correction and radiometric correction, but has not been orthorectified using a digital elevation model (DEM).
We compile a total of 13 image sequences to create this dataset, with sequence lengths ranging from 2 to 5 frames and time spans from several minutes to a few days.
From these sequences, multiple ship tracks are extracted.
The 13 sequences include 18 SAR images and 25 optical images, amounting to a total of 43 frames.
Most sequences contain both modalities.
We manually annotate the rectangular bounding boxes around the ships and crop them out.
Subsequently, we manually identify and associate the targets to generate a complete image sequence for each ship.
\begin{figure}[t]
\centering
\includegraphics[width=3.2in]{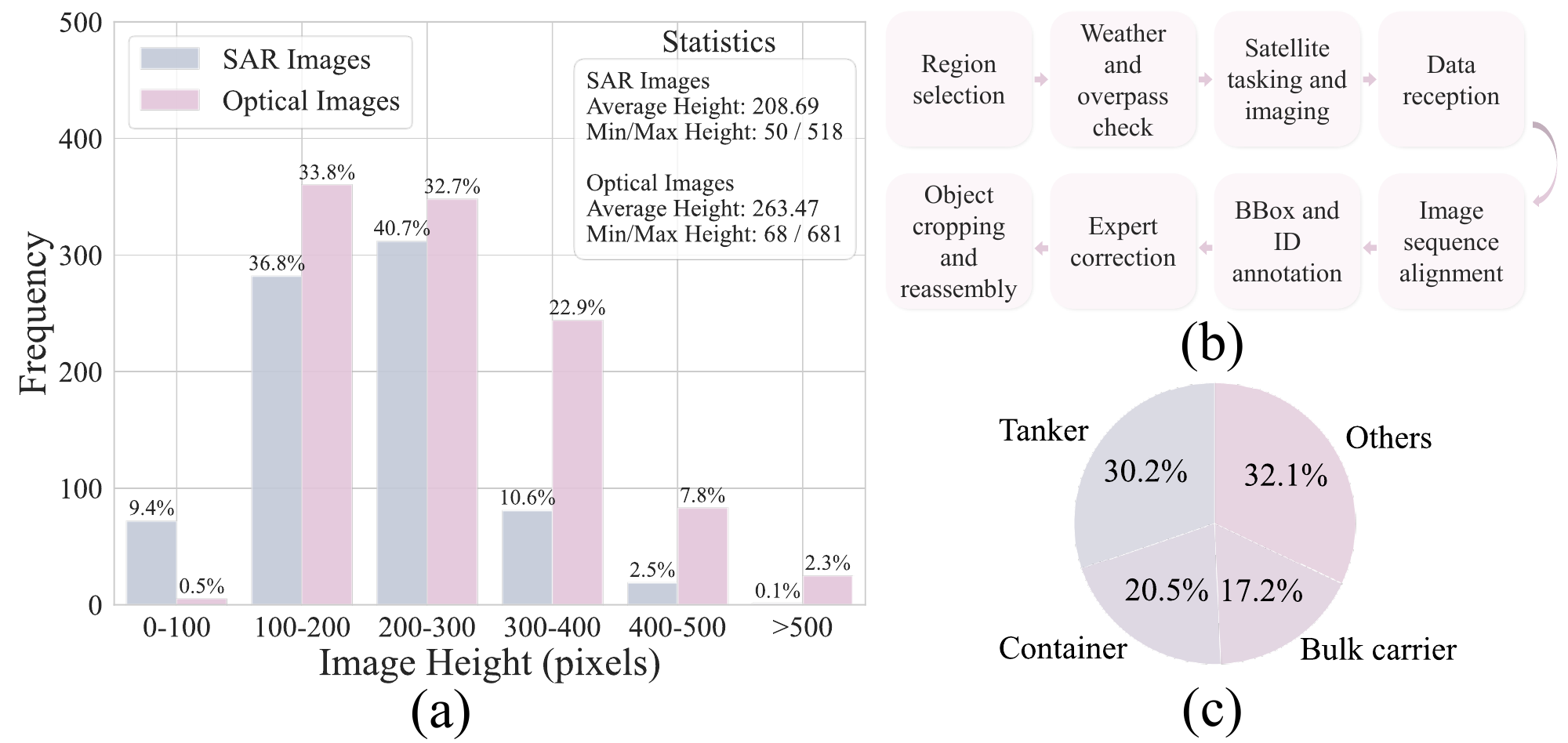}
\hfil
\vspace{-1.6ex}\caption{(a) Distribution of ship slice heights. (b) Dataset construction process. (c) Distribution of ship categories.\vspace{-1ex}}
\label{statistics}
\end{figure}
\begin{table}[t]
\setlength\tabcolsep{7pt}
\centering
\begin{tabular}{ccccc}
\toprule
\multirow{2.5}{*}{Split} & \multirow{2.5}{*}{Tracks} & \multicolumn{3}{c}{Images} \\
\cmidrule(lr){3-5}
\multirow{2}{*}{} & \multirow{2}{*}{} & Optical & SAR & All \\
\midrule
Train & 361 & 574 & 489 & 1063 \\
Query & 88 & 88 & 88 & 176 \\
Gallery & 88 (+163) & 403 & 190 & 593 \\
\midrule
All & 449 (+163) & 1065 & 767 & 1832 \\
\bottomrule
\end{tabular}
\caption{Detailed statistics for HOSS ReID dataset. ``(+163)" indicates there are 163 distractor objects added to the gallery.\vspace{-2.3ex}}
\label{tab:table1}
\end{table}
\begin{figure*}[t]
\centering
\includegraphics[width=5.9in]{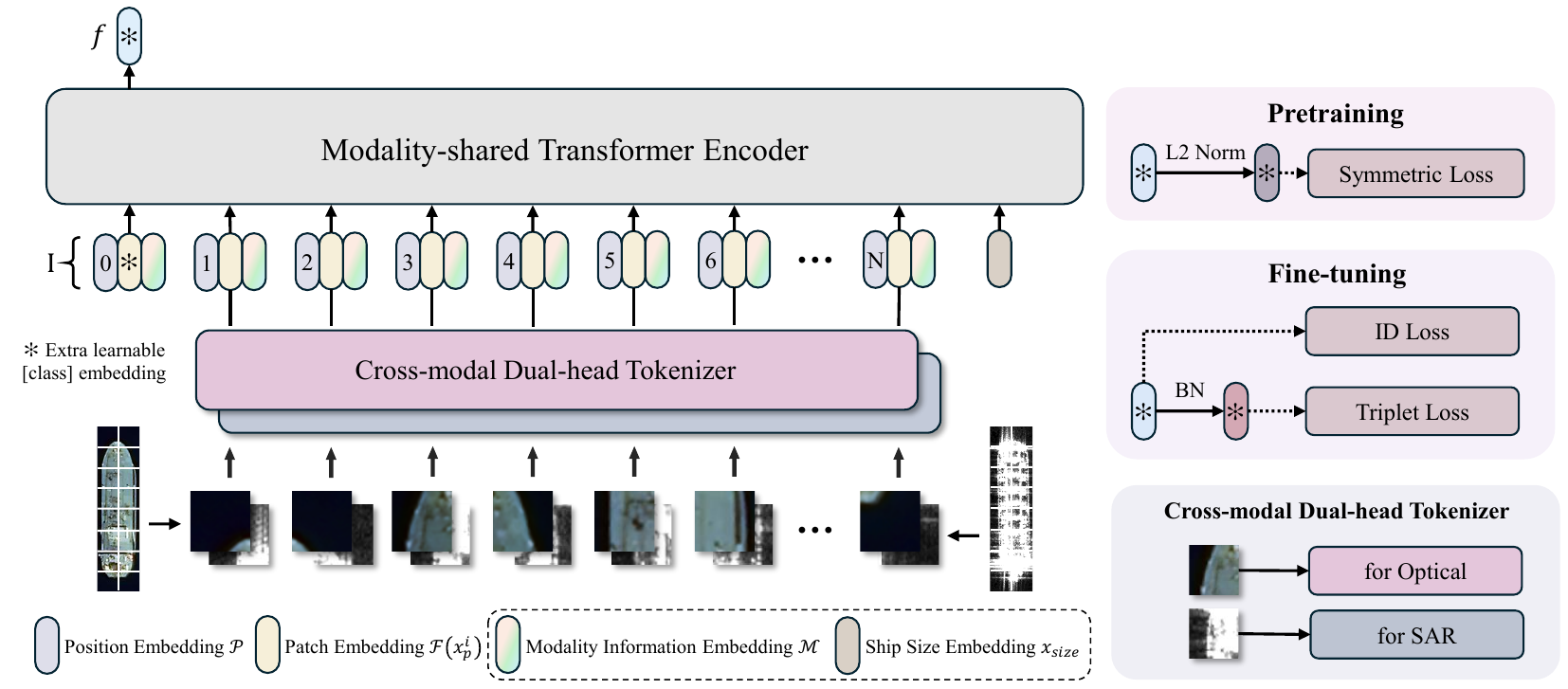}
\hfil
\caption{Framework of proposed TransOSS. Its overall structure is based on ViT, but it incorporates a dual-head tokenizer, modality information embeddings, and ship size embeddings. The network processes one image at a time, utilizing different tokenizers for different modalities but sharing the same transformer encoder for feature extraction. The ship size embedding is treated as a separate vector, placed alongside and following the other image embeddings.\vspace{-2.6ex}}
\label{Network}
\end{figure*}
\vspace{-2ex}Our dataset format is modeled after the Market-1501 \cite{ref39} dataset, with filenames containing the object ID, sequence ID, camera ID, and modality.
The sequence ID specifies which of the 13 sequences the image belongs to, while the camera ID indicates which satellite captured the image within that sequence.
All image files are in ``TIF" format, with optical images formatted as 3-channel 8-bit RGB and SAR images as single-channel 32-bit floating-point to preserve more detailed information.
Some statistical information of the dataset is presented in Figure \ref{statistics}: the length distribution of images (reflecting the size distribution of ships), the dataset construction process, and the general category distribution.

We divide the images into training and validation sets, with specific image and object counts detailed in Table \ref{tab:table1}.
To enhance the dataset's complexity and better simulate real-world scenarios, we include 163 distractor objects in the gallery that do not belong to the query, labeled as ``-1".
These distractor images share the same origin as other images but correspond to no specific queries, serving to expand the gallery scale and simulate real-world scenarios containing abundant irrelevant objects.
Our dataset features queries consisting of both optical and SAR images, each accounting for 50\%.
However, the number of each modality per object in the gallery is arbitrary.
Similarly, in the training set, the distribution of modalities per object is also arbitrary, with a small proportion of objects represented by images of only one modality.

\section{Methodology}
\label{sec:method}
We propose a transformer-based optical-SAR ship ReID method (TransOSS).
The proposed network is capable of extracting features from both optical and SAR images, mapping these cross-modality images into a shared feature space with a dimensionality of $D$ ($D=768$ in our model).
In this unified feature space, we employ the Euclidean distance as the metric to compute the similarity between pairs of images, thereby constructing a distance matrix.
Based on this distance matrix, we can further determine the object corresponding to the query image, achieving cross-modal ship ReID tasks.

\subsection{Overall Architecture of TransOSS}
\label{sec:4.1}
The overall network structure is illustrated in Figure \ref{Network}.
It is a fully transformer-based architecture.
The input image is splited into $N$ fixed-size patches, which are passed through a linear layer and then summed with both position embeddings $\mathcal{P}$ and modality information embeddings $\mathcal{M}$.
Following the ViT architecture, we introduce a learnable [class] embedding $x_{cls}$, whose output serves as the global feature representation, $f$.
Additionally, we include a embedding $x_{size}$ representing the ship's actual size.
Consequently, the complete sequence entering the modality-shared transformer encoder consists of the following components:
\begin{equation}
\mathrm{I}=\mathcal{P}+\left[x_{cls};\mathcal{F}\left(x_p^1\right);\cdots;\mathcal{F}\left(x_p^N\right)\right]+\mathcal{M};x_{size}
\label{equ:equation1}
\end{equation}
where $\mathcal{F}$ is the cross-modal dual-head tokenizer, and $x_p^i$ is $i$-th image patch.

{\bf Cross-modal dual-head tokenizer.}
In cross-modal ReID tasks, researchers often employ two separate backbones for feature extraction \cite{ref11, ref12, ref28, ref40}.
However, we adopt a more lightweight approach.
We utilize separate tokenizers for optical and SAR images, employing a simple linear layer.
This approach ensures that images from different modalities are mapped independently, effectively suppressing irrelevant information at an early stage while adding only a minimal number of parameters.
Leveraging the powerful feature representation capabilities of ViT, we first embed the images and then apply a modality-shared transformer encoder for feature extraction.
This process facilitates mapping the features into a common space, and simplifying the training process.

\subsection{Auxiliary Information Learning}
\label{sec:4.2}
To further boost the network's performance, we introduce auxiliary information for the network to learn, without modifying its structure.

{\bf Modality information embeddings.}
After obtaining the image embeddings through the dual-head tokenizer, the features extracted by the model may still exhibit modality-bias, as both modalities share a single transformer encoder.
The model's perception of modalities helps generate modality-invariant features.
Therefore, we introduce modality information embeddings inspired by the approach in \cite{ref47, ref10}.
As defined in Eq. \ref{equ:equation1}, the modality embedding $\mathcal{M}$ is directly added to the image embeddings and treated as a learnable parameter, initialized with a normal distribution.
Each modality is assigned a unique embedding, and all image patches corresponding to a particular modality share the same embedding.
Specifically, before the model training begins, we initialize $\mathcal{M} \in \mathbb{R}^{2 \times D}$.
The embeddings for optical and SAR data correspond to $\mathcal{M}\left[0\right]$ and $\mathcal{M}\left[1\right]$ respectively.
For the $i$-th input to the encoder, the formulation is as follows:
\begin{equation}
I_i=\mathcal{P}_i+\mathcal{F}\left(x_p^i\right)+\lambda\mathcal{M}\left[x\right],\ x=\left\{\begin{matrix}
 0 & if\ optical\\
 1 & if\ SAR
\end{matrix}\right.
\end{equation}
where $\mathcal{P}_i$ is the $i$-th position embedding, and $\lambda$ is a hyperparameter that balances the weight.

{\bf Ship size embeddings.}
Unlike conventional images, remote sensing imagery enables direct measurement of ground object dimensions, allowing for approximate calculation of ship length and width based on the GSD.
However, most current ReID models resize input images to a uniform size, leading to a loss of crucial size-related information \cite{ref49}.
In the context of ship ReID, object size and aspect ratio are critical features that can be leveraged to enhance identification accuracy.
So, we incorporate ship size information via a linear layer and feed it into the encoder alongside the other image embeddings:
\begin{equation}
x_{size}=Linear\left(i_w,i_h,i_{w/h}\right)
\label{equ:equation3}
\end{equation}
where $i_w$ and $i_h$ are the normalized and standardized width and length of the ship, and $i_{w/h}$ is the ship's aspect ratio.
$Linear\left(\cdot\right)$ is a fully connected layer with an input dimension of 3 and an output dimension of $D$. 
The calculation of the ship’s length and width is performed by directly multiplying the image’s dimensions by GSD, with each image being processed independently.
However, since the image boundaries do not tightly align with the ship's edges, and the images have not undergone orthorectification or other precise corrections, the measurements can only approximate the ship’s size and proportions.
Despite this, it provides sufficient information for the model to efficiently distinguish ships with significant scale differences.

\subsection{Two-Stage Training Approach}
\label{sec:4.3}
Given the substantial differences between optical and SAR images, coupled with the limited pre-training of existing backbones on datasets containing SAR images, we propose a two-stage training approach.
This method allows the model to leverage a large amount of training data and effectively learn to extract modality-invariant features from both image types.

{\bf Pretraining.}
While acquiring multimodal remote sensing images (optical and SAR) of the same object is challenging, previous researchers have developed several optical-SAR paired datasets for tasks such as image fusion and instance segmentation \cite{ref41, ref42, ref43, ref44}.
These optical-SAR pairs can be utilized to pre-train our network, allowing it to learn shared feature extraction across both modalities.
For this purpose, we use the SEN1-2 \cite{ref41} and DFC23 \cite{ref42} datasets.
SEN1-2 offers a broader variety of scenes with a larger number of images, while DFC23 provides higher spatial resolution.
\begin{figure}[t]
\centering
\includegraphics[width=2.6in]{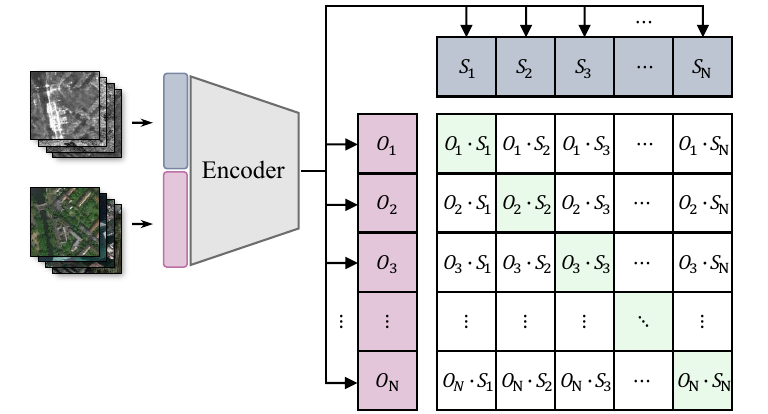}
\hfil
\caption{TransOSS pre-training approach. This method involves inputting multiple pairs of optical and SAR images into the network. The objective is to minimize the distance between paired images using a symmetric loss function.\vspace{-2.5ex}}
\label{Pretrain}
\end{figure}

We adopt a training approach inspired by methods like CLIP \cite{ref45, ref46}, as illustrated in Figure \ref{Network} and Figure \ref{Pretrain}, which leverages contrastive learning with paired optical-SAR images. This approach effectively aligns features from the two modalities by maximizing similarity within corresponding pairs while minimizing it across non-corresponding pairs.
The network is optimized using a symmetric cross-entropy loss, defined as follows:
\begin{equation}
\mathcal{L}_{\text{pre}}=\frac{ce\left[\left(O \cdot S^T\right)\times\sqrt\tau\right]+ce\left[\left(S \cdot O^T\right)\times\sqrt\tau\right]}{2}
\label{equ:equation2}
\end{equation}
where $ce$ denotes the cross-entropy loss, while $O$ and $S$ represent the L2-normalized features of the optical and SAR images, respectively. The variable $\tau$ is a learnable hyperparameter.
In this step, we introduce modality information embeddings, but we do not include ship size embeddings.

{\bf Fine-tuning.}
During network fine-tuning on our HOSS ReID dataset, we employ the widely used ID loss and triplet loss, with the ID loss formulated as cross-entropy loss.
A classification head is added to the network, enabling it to learn to classify ships in the training set, with each ID representing a distinct category.
We adopt a triplet data loading format, where each sample group consists of an anchor $a$, a positive sample $p$, and a negative sample $n$.
The triplet loss functions by constraining the model to bring the features of the same object closer together, while pushing the features of different objects further apart.
The loss is calculated as follows:
\begin{equation}
\mathcal{L}_{\text{triplet}} = \max\left( d(a, p) - d(a, n) + \text{margin}, 0 \right)
\end{equation}
where $d\left(\cdot\right)$ represents the Euclidean distance, and $margin$ is a pre-defined safety margin.
\section{Experiments}
\label{sec:experi}

\begin{table*}[t]
\centering
\begin{tabular}{cccccccccccccc}
\toprule
\multirow{2.5}{*}{Backbone} & \multirow{2.5}{*}{Method} & \multicolumn{4}{c}{All} & \multicolumn{4}{c}{Optical to SAR} & \multicolumn{4}{c}{SAR to Optical}  \\
\cmidrule(lr){3-6}\cmidrule(lr){7-10}\cmidrule(lr){11-14}
{} & {} & mAP & R1 & R5 & R10 & mAP & R1 & R5 & R10 & mAP & R1 & R5 & R10 \\
\hline\hline
\multirow{5}{*}{ResNet} & CM-NAS \cite{ref50} & 30.7 & 46.0 & 54.6 & 57.4 & 8.2 & 1.5 & 10.8 & 21.5 & 7.6 & 4.5 & 11.9 & 19.4 \\
{} & LbA \cite{ref28} & 33.0 & 48.3 & 59.7 & 62.5 & 11.9 & 4.6 & 23.1 & 41.5 & 8.5 & 6.0 & 14.9 & 22.4 \\
{} & Hc-Tri \cite{ref56} & 34.0 & 47.2 & 54.6 & 59.7 & 11.1 & 6.2 & 15.4 & 24.6 & 10.9 & 7.5 & 20.9 & 29.9 \\
{} & AGW \cite{ref49} & 43.6 & 57.4 & 64.2 & 68.8 & 17.2 & 7.7 & 29.2 & 38.5 & 21.1 & 14.9 & 34.3 & 46.3 \\
{} & DEEN \cite{ref30} & 43.8 & 58.5 & 64.2 & 66.5 & 31.3 & 21.5 & 44.6 & 60.0 & 27.4 & 22.4 & 40.3 & 53.7 \\
{} & MCJA \cite{ref62} & 47.1 & 59.1 & 67.9 & 73.0 & 18.6 & 10.8 & 27.7 & 38.5 & 19.7 & 14.9 & 28.3 & 43.3 \\
\hline
\multirow{6}{*}{Transformer} & SOLIDER \cite{ref63} & 38.2 & 50.6 & 63.1 & 69.9 & 23.1 & 12.3 & 38.5 & 52.3 & 14.6 & 10.4 & 16.4 & 31.3 \\
{} & ViT-base \cite{ref17} & 43.0 & 56.2 & 64.8 & 69.9 & 21.5 & 12.3 & 33.8 & 55.4 & 17.9 & 10.4 & 25.4 & 32.8 \\
{} & DeiT-base \cite{ref64} & 47.2 & 58.1 & 69.6 & 74.1 & 25.9 & 16.1 & 36.7 & 59.1 & 26.1 & 20.3 & 35.7 & 52.0 \\
{} & TransReID \cite{ref47} & 48.1 & 60.8 & 69.3 & 73.9 & 27.3 & 18.5 & 40.0 & 58.5 & 20.9 & 11.9 & 34.3 & 43.3 \\
{} & VersReID \cite{ref65} & 49.3 & 59.7 & 70.5 & 78.4 & 25.7 & 13.8 & 40.0 & 61.5 & 27.7 & 17.9 & 44.8 & 61.2 \\
{} & TransOSS* & 49.4 & 61.9 & 71.0 & 78.4 & 30.2 & 16.9 & 49.2 & 63.1 & 29.1 & 20.9 & 49.3 & 64.2 \\
{} & TransOSS & \pmb{57.4} & \pmb{65.9} & \pmb{79.5} & \pmb{85.8} & \pmb{48.9} & \pmb{33.8} & \pmb{67.7} & \pmb{80.0} & \pmb{38.7} & \pmb{29.9} & \pmb{59.7} & \pmb{71.6} \\
\bottomrule
\end{tabular}
\caption{Comparisons of CMC (\%) and mAP (\%) performances on the HOSS ReID dataset. The star * indicates TransOSS without pre-training on optical-SAR image pairs. ``Optical to SAR" means that queries from the optical modality are searched only within the SAR modality gallery, and ``SAR to Optical" has a similar meaning.\vspace{-2ex}}
\label{tab:table2}
\end{table*}
\subsection{Implementation Details}
We apply several data augmentation techniques, including random horizontal flipping, cropping, and erasing \cite{ref48}.
During pre-training, we select a subset of images from the SEN1-2 dataset and all images from the DFC23 dataset, cropping them to $128\times256$, resulting in approximately 56K image pairs.
We use a batch size of 512 and train the model for 60 epochs using stochastic gradient descent (SGD) with a learning rate of 3e-3.
The initial model weights are derived from the ViT-base model pre-trained on ImageNet.
For fine-tuning on the HOSS ReID dataset, we resize all images to $128\times256$ (which performs best in the experiments), reducing the batch size to 32, and train for 200 epochs, again using SGD but with a lower learning rate of 5e-4.

Larger batch size gives better results in pre-training, so we utilize 4 NVIDIA A100 40GB GPUs to meet the memory requirements, and the algorithm is implemented using the PyTorch framework.
For evaluation, we employ standard metrics commonly used in ReID tasks, including Cumulative Matching Characteristic (CMC) curves and mean Average Precision (mAP), to assess the algorithm's performance.

\subsection{Model Comparisons and Analysis}
We compare our method against several state-of-the-art open-source ReID algorithms, retraining each on the HOSS ReID dataset. 
The comparison include transformer-based models (SOLIDER, ViT, DeiT, TransReID, VersReID) as well as ResNet-based methods tailored for infrared-visible cross-modal ReID (CM-NAS, LbA, Hc-Tri, AGW, DEEN, MCJA), where DeiT and ViT is the implementation from reference \cite{ref47}.
For consistency, the input size for all networks is same, and the original hyperparameters are retained.
Each model is trained for a sufficient number of epochs to ensure fair evaluation.

The detailed experimental results are presented in Table \ref{tab:table2}.
ViT-based methods consistently outperform ResNet-based methods on the HOSS ReID dataset.
Despite being designed for cross-modal ReID, the ResNet-based models, with modules specifically tailored for infrared-visible ReID, fail to perform optimally when applied to optical-SAR ReID.
Among these, the MCJA method for cross-modal person ReID achieves the highest performance.
In contrast, ViT-based models show significantly better mAP and CMC metrics.
Without any modifications, ViT-base outperforms most ResNet-based methods.
TransReID and VersReID, enhanced versions of ViT, further improve performance.
Our proposed method, TransOSS, which includes task-specific optimizations, shows strong results even without pre-training on optical-SAR image pairs, leveraging only ViT weights pre-trained on ImageNet.
With two-stage training approach, the mAP improves to 57.4\%, far surpassing all other methods, while rank-1 accuracy reaches 65.9\%, the highest among all approaches.

\subsection{Cross-modal experiments}
\begin{figure}[b]
\vspace{-2ex}
\centering
\includegraphics[width=2.8in]{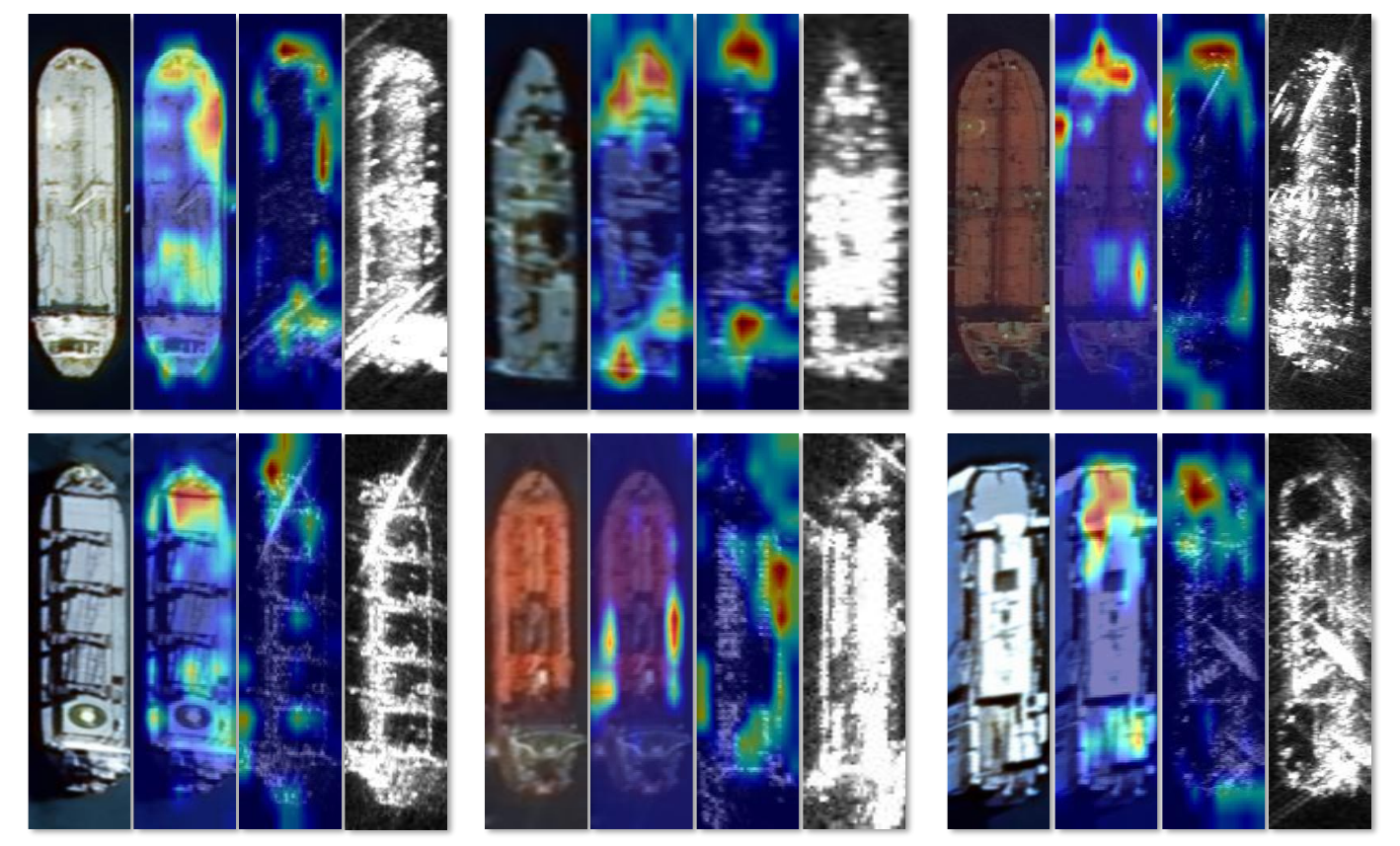}
\hfil
\caption{Grad-CAM visualization of attention maps. For each set of images, from left to right, they are: optical modality image, optical modality image attention map, SAR modality image attention map, and SAR modality image.\vspace{-2ex}}
\label{Feature}
\end{figure}
Achieving accurate cross-modal ReID is the most challenging aspect of this task, particularly when performing ReID from the more information-limited SAR modality to the optical modality.
To evaluate this, we adjust the validation set to include only cross-modal cases, conducting ReID experiments in both optical-to-SAR and SAR-to-optical directions.
The results are presented in Table \ref{tab:table2}.
Aside from the outstanding performance of DEEN, most existing methods achieve only marginal accuracy in cross-modal ReID tasks, suggesting that the infrared-visible ReID approach is difficult to apply to this scenario.
Our proposed method significantly improves performance, achieving an mAP of 48.9\% and a rank-1 accuracy of 33.8\% in the optical-to-SAR scenario.
In the SAR-to-optical scenario, TransOSS performs slightly lower due to the inherent feature scarcity in SAR images, which makes using SAR as the query more challenging for ReID.
Nevertheless, it still significantly outperforms the baseline.
These results validate the effectiveness of our approach, though there remains substantial room for further accuracy improvements.

To further validate the effectiveness of our method for cross-modal ReID, we employed Grad-CAM \cite{ref51} to visualize the attention maps of TransOSS, as shown in Figure \ref{Feature}.
They reveal that the model focuses on similar local areas and ship contours across both modalities, demonstrating that it has learned to capture both modality-invariant details and global features, such as the ship’s shape.

\subsection{Ablation Study}
\begin{table}[b]
\vspace{-3ex}
\centering
\begin{tabular}{cccc|cc}
\toprule
CDT & Pretraining & MIE & SSE & mAP & R1 \\
\hline\hline
\textcolor{lightgray}{\ding{55}} & \textcolor{lightgray}{\ding{55}} & \textcolor{lightgray}{\ding{55}} & \textcolor{lightgray}{\ding{55}} & 43.0 & 56.2 \\
\ding{51} & \textcolor{lightgray}{\ding{55}} & \textcolor{lightgray}{\ding{55}} & \textcolor{lightgray}{\ding{55}} & 44.9 & 58.0 \\
\textcolor{lightgray}{\ding{55}} & \ding{51} & \textcolor{lightgray}{\ding{55}} & \textcolor{lightgray}{\ding{55}} & 52.3 & 62.7 \\
\ding{51} & \ding{51} & \textcolor{lightgray}{\ding{55}} & \textcolor{lightgray}{\ding{55}} & 53.2 & 63.1 \\
\hline
\ding{51} & \ding{51} & \ding{51} & \textcolor{lightgray}{\ding{55}} & 53.8 & 64.2 \\
\ding{51} & \ding{51} & \textcolor{lightgray}{\ding{55}} & \ding{51} & 55.0 & 64.2 \\
\ding{51} & \ding{51} & \ding{51} & \ding{51} & \pmb{57.4} & \pmb{65.9} \\
\bottomrule
\end{tabular}
\caption{The ablation study of TransOSS. ``CDT" represents cross-modal dual-head tokenizer, ``MIE" represents modality information embeddings, and ``SSE" represents ship size embeddings in the table.}
\label{tab:table3}
\end{table}
We conduct ablation experiments to validate the effectiveness of the proposed optical and SAR feature alignment approach and the auxiliary information learning method.
The complete experimental results are presented in Table \ref{tab:table3}.
ViT-base is used as the baseline.

{\bf Effect of optical and SAR feature alignment.}
When the cross-modal dual-head tokenizer is introduced, with both heads initialized using ImageNet pre-trained weights without specific training on SAR data, we observe a 4.4\% improvement in mAP.
This is achieved by embedding images into a shared feature space while maintaining separation for highly differentiated images.
When both baseline and our method undergo pre-training, the dual-head tokenizer also results in a 1.9 percentage points mAP improvement.
After pre-training, ViT-base shows a significant increase, reaching 52.3\% mAP and 62.7\% rank-1 accuracy.
These results highlight that the utilization of large-scale datasets is essential.
With our proposed pre-training approach, external data can be effectively leveraged, resulting in substantial performance gains.

{\bf Effect of auxiliary information learning.}
The experimental results demonstrate that operating on embeddings and incorporating auxiliary information yields performance improvements with minimal impact on network computational complexity.
The introduction of modality information embeddings improves rank-1 accuracy by 1.1 percentage points, while ship size embeddings lead to a significant 1.8 percentage points increase in mAP.
With all proposed enhancements, TransOSS achieves a remarkable improvement of 33.5\% in mAP and 17.3\% in rank-1 accuracy compared to the baseline, highlighting the effectiveness of our design.

\subsection{Discussions}
Unlike conventional images, remote sensing imagery offers physically meaningful information, such as geographic coordinates and the physical size of objects, and SAR images further reflect physical characteristics, such as specific structures that generate scattering points, providing valuable additional information that can be utilized.
Alternatively, self-supervised or unsupervised methods \cite{ref57, ref58, ref59} should be explored, alongside the incorporation of additional data augmentation techniques \cite{ref60, ref61}, which are particularly effective in data-scarce scenarios.
Future work could also incorporate modalities like text and multispectral imagery to enhance applicability across diverse scenarios.

\section{Conclusion}
\label{sec:conclu}
We decompose the remote sensing ship tracking task into LEO-based object detection, ReID, and trajectory generation, with particular investigation focused on the ReID component.
We introduce the HOSS ReID dataset and the TransOSS method, with a focus on cross-modal ship ReID to leverage a larger satellite network, shorten revisit times, and mitigate the effects of weather.
Our dataset is meticulously designed to simulate real-world scenarios using programmed satellite imaging, offering a challenging benchmark for evaluating methods' ability to link the same object within and across modalities in large datasets.
TransOSS capitalizes on the unique challenges of optical-SAR cross-modality by optimizing its network architecture, pretraining on large-scale datasets through contrastive learning, and incorporating auxiliary information to enhance the model's capability to extract shared features between optical and SAR images.
Experimental results show that our method achieves a substantial improvement in accuracy.

\section*{Acknowledgement}
This work was supported by the Key Deployment Program of the Chinese Academy of Sciences (Grant No. KGFZD-145-23-18).

{
    \small
    \bibliographystyle{ieeenat_fullname}
    \bibliography{main}
}

\end{document}